\documentclass[review]{elsarticle}

\usepackage{lineno,hyperref}
\usepackage[linesnumbered,ruled,vlined]{algorithm2e}
\usepackage{algpseudocode}  
\usepackage{amsmath}  
\usepackage{color}
\usepackage{graphicx}
\usepackage{booktabs}
\usepackage{pgf}
\usepackage{import}
\usepackage{float}
\journal{Journal of \LaTeX\ Templates}

\begin{document}

\begin{frontmatter}

\title{Unveiling personnel movement in a larger indoor area with a non-overlapping multi-camera system}


\author[mymainaddress,mysecondaryaddress]{Ping Zhang}

\author[mymainaddress,mysecondaryaddress]{Zhenxiang Tao}
\author[mymainaddress,mysecondaryaddress]{Wenjie Yang}
\author[mymainaddress,mysecondaryaddress]{Minze Chen}
\author[mymainaddress,mysecondaryaddress]{Shan Ding}

\author[mythirdaddress]{Xiaodong Liu}

\author[mymainaddress,mysecondaryaddress]{Rui Yang\corref{mycorrespondingauthor}}
\cortext[mycorrespondingauthor]{Corresponding author}
\ead{ryang@tsinghua.edu.cn}

\author[mymainaddress,mysecondaryaddress]{Hui Zhang}

\address[mymainaddress]{Department of Engineering Physics, Tsinghua University, Beijing 100084, China}
\address[mysecondaryaddress]{Institute of Public Safety Research, Tsinghua University, Beijing 100084, China}
\address[mythirdaddress]{Public Order School, People's Public Security University of China, 100038, Beijing, China}

\begin{abstract}
Surveillance cameras are widely applied for indoor occupancy measurement and human movement perception, which benefit for building energy management and social security. To address the challenges of limited view angle of single camera as well as lacking of inter-camera collaboration, this study presents a non-overlapping multi-camera system to enlarge the surveillance area and devotes to retrieve the same person appeared from different camera views. The system is deployed in an office building and four-day videos are collected. By training a deep convolutional neural network, the proposed system first extracts the appearance feature embeddings of each personal image, which detected from different cameras, for similarity comparison. Then, a stochastic inter-camera transition matrix is associated with appearance feature for further improving the person re-identification ranking results. Finally, a noise-suppression explanation is given for analyzing the matching improvements. This paper expands the scope of indoor movement perception based on non-overlapping multiple cameras and improves the accuracy of pedestrian re-identification without introducing additional types of sensors.
\end{abstract}


\end{frontmatter}

\section{Introduction}

For intelligent buildings, accurate indoor occupancy measurement and individual moving information acquisition are indispensable prerequisites to improve energy efficiency and security management. Related applications, like customer counting in shopping malls, space utilization analysis in multi-story buildings, and passenger behavior analytic in transportation hubs, all require indoor personal location and movement information. Considering the easy installation and decreased costs of surveillance cameras, an increasing number of research have been devoted to treating the video processing as an active perception method for indoor personal sensing \cite{Sun2020,Saha2019,Rueda2020}. Meanwhile, computer vision techniques, including person detection, recognition, and re-identification are applied to indoor surveillance video analysis for occupant detection, zonal person counting, and movement pattern analyzing. 

Considering the dynamic characteristics of indoor personal movements, when people moving inside buildings, they will be captured by multiple cameras which always have no field-of-view overlap. To figure out the individuals’ fully moving patterns, it is necessary to associate the incomplete information captured by each single camera with cross-view person re-identification techniques. Person re-identification can be treated as retrieving the same people who appeared in other cameras by comparing appearance feature similarity. However, the main encountered challenges include: (1) people dressing looks alike which bring difficulties to distinguish different individuals; (2) the camera view angle and personal pose changed when person moving among different indoor surveillance spaces; (3) the complete track acquisition are impeded when people moving across the blind area between two neighboring cameras with no filed-of-view overlap.

To conquer the difficulties mentioned above, this study first deploys multiple non-overlapping cameras in different locations of an office building. Correspondingly, a graph-based camera linkage model is established to capture indoor personal movement over a specific floor area. Based on the detected personal images from different camera videos, distinctive appearance feature embeddings are extracted by a deep convolutional neural network for similarity comparison. Furthermore, statistical indoor transition patterns, including stochastic transition choice matrix, are coupled with appearance features for improving the cross-camera person matching accuracy. The main contributions to the community include:

\begin{itemize}
\item A multi-camera video-based person re-identification framework is proposed for enlarging surveillance areas and unveiling indoor person moving patterns in larger scale. 
\item Stochastic indoor personal transition matrix is integrated with appearance features for improving cross-camera person re-identification accuracy.
\item A noise suppression-based theoretical analysis is given for explaining the accuracy improvements.
\end{itemize}

\section{Related Works}
Surveillance cameras can be treated as visual sensors for indoor personal monitoring and sensing \cite{Benezeth2011,Yang2016,Wang2013}. The uses of video surveillance to analyze the distribution of indoor occupants are passive detection scheme, which do not requires the cooperation of pedestrians, and has advantage in reflecting their daily movement characteristics. By deploying ceiling-mounted Kinect camera at the room entrance areas, Petersen et al. \cite{Petersen2016} presented image processing techniques to detect, track and count indoor occupants. To avoid the body parts occlusion problem happens in camera views, Zou et al. \cite{Zou2017} and Chen et al. \cite{Chen2009} proposed a head detection algorithm for measuring indoor occupancy. Gao et al. \cite{Gao2016} further applied the head detection-based people counting method to crowded classroom surveillance environment. Combing with computer vision-based tracking method, Kuipers et al. \cite{Kuipers2014} deployed four cameras in an atrium and carried out several tests to study the dynamics of mobility. 

However, the above-mentioned research mainly focuses on single distinct camera cases inside partial indoor areas without taking the information fusion between neighboring cameras into account. When confronted with larger indoor area which cannot fully covered by a single camera, these methods will fail to restore the complete personal moving activities. To enlarge the coverage area, several researches deploy PTZ camera for indoor occupant counting and monitoring \cite{Shih2014,ding2012collaborative}. However, the physical limitation of the complicated indoor structure impedes the widely usage of PTZ cameras. Instead, researchers are devoted to deploying camera networks, which consists of multiple non-overlapping cameras to maximize the coverage of personal movements. To measure the indoor occupancy, Liu et al. \cite{Liu2013} proposed a dynamic Bayesian network-based method to combine the detection results from multiple vision sensors, which deployed at the room entrance and interior. Camps et al. \cite{Camps2017} deployed a human re-identification system inside an airport, which composed of three cameras. In addition, several public indoor video data sets are released for multi-camera person re-identification and tracking. Figueira et al. \cite{Figueira2015} deployed a heterogeneous camera network in an indoor office, which consists of 13 cameras distributed over three floors. Bialkowski et al. \cite{Bialkowski2012} proposed a multi-camera database, which consists of 8 cameras inside building environment, for the task of person re-identification in surveillance networks. Marroquin et al. \cite{Marroquin2019} created WiseNET dataset for people detection and tracking, which composed of 6 indoor cameras in a building floor with contextual information and annotations. By deploying 15 synchronized cameras mounted in corridor and junction, Styles et al. \cite{Styles2020} devote to predicting the personal location within camera networks.

With the ever increasing of big visual data and the rising of computing power, much research efforts are now dedicated to performing deep learning based methods for extracting more distinctive appearance feature embeddings, which applied for learning feature distance between image pairs \cite{Bedagkar-Gala2014, Zheng2016, Karanam2019, Leng2020, Ye2020}. To fill the gap of the discontinuity of pedestrian data across non-overlapping cameras, indoor spatial-temporal information, including camera topology, cross-camera transition choice, and travel time distributions are exploited to improve the accuracy of person re-identification results \cite{Chen2011,Chen2014,Javed2008,Wang2019}. In this work, we propose an indoor camera linkage model, and formulate the movement of pedestrian in buildings as the information flow among camera nodes. By training a deep convolutional neural network-based feature extractor, and leveraging building information as well as the transition matrix of pedestrian movement, the cross-camera pedestrian matching is realized.

\section{Experiment and Methodology}

\subsection{Framework and modules}
\begin{figure}[h]
	\centering
	\includegraphics[scale=0.7]{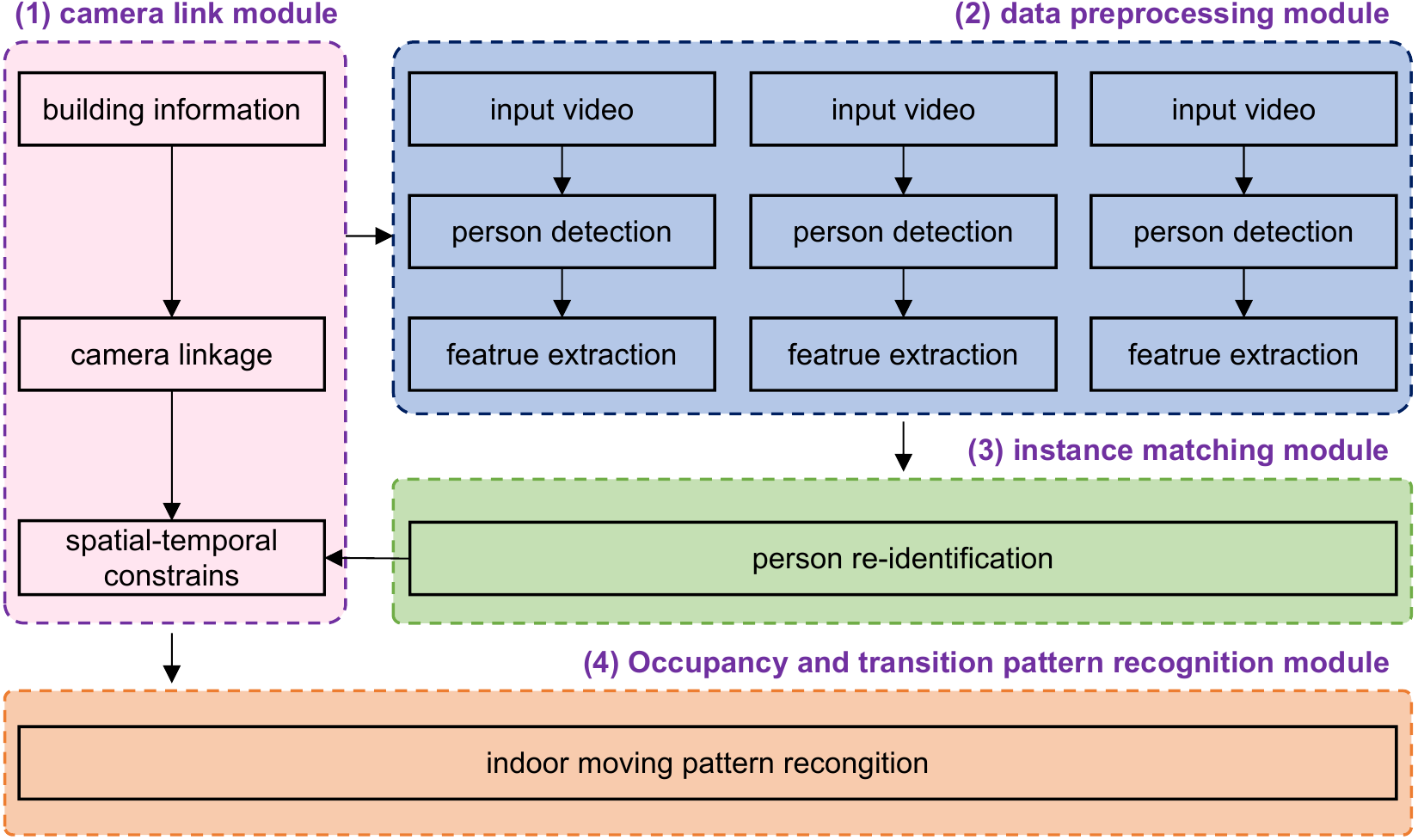}
	\caption{Framework and pipeline of this study}
	\label{fig:framework}
\end{figure}

An indoor multiple camera system, which composed of four modules, is deployed for capturing personal movement in larger indoor area, as shown in Figure \ref{fig:framework}. First, integrating with the prior knowledge about building structure and surveillance camera placement, topological camera linkage is established, and raw videos are collected from these non-overlapping cameras. Secondly, by using an off-the-shelf person detection approach, all of the person who appeared in these cameras are detected. The detected images are then cropped from each video and annotated with global unique labels. By using the supervised learning method, a deep convolutional neural network (ConvNets) is trained to extract appearance features for similarity comparison. Furthermore, by ranking the similarity distance, same individual who appeared at different cameras are matched. Finally, we devoted to recognizing indoor personal moving patterns from the person re-identification results combined with spatial-temporal constrains.

\subsection{Data preparation}
\begin{figure}[h]
	\centering
	\includegraphics[scale=0.7]{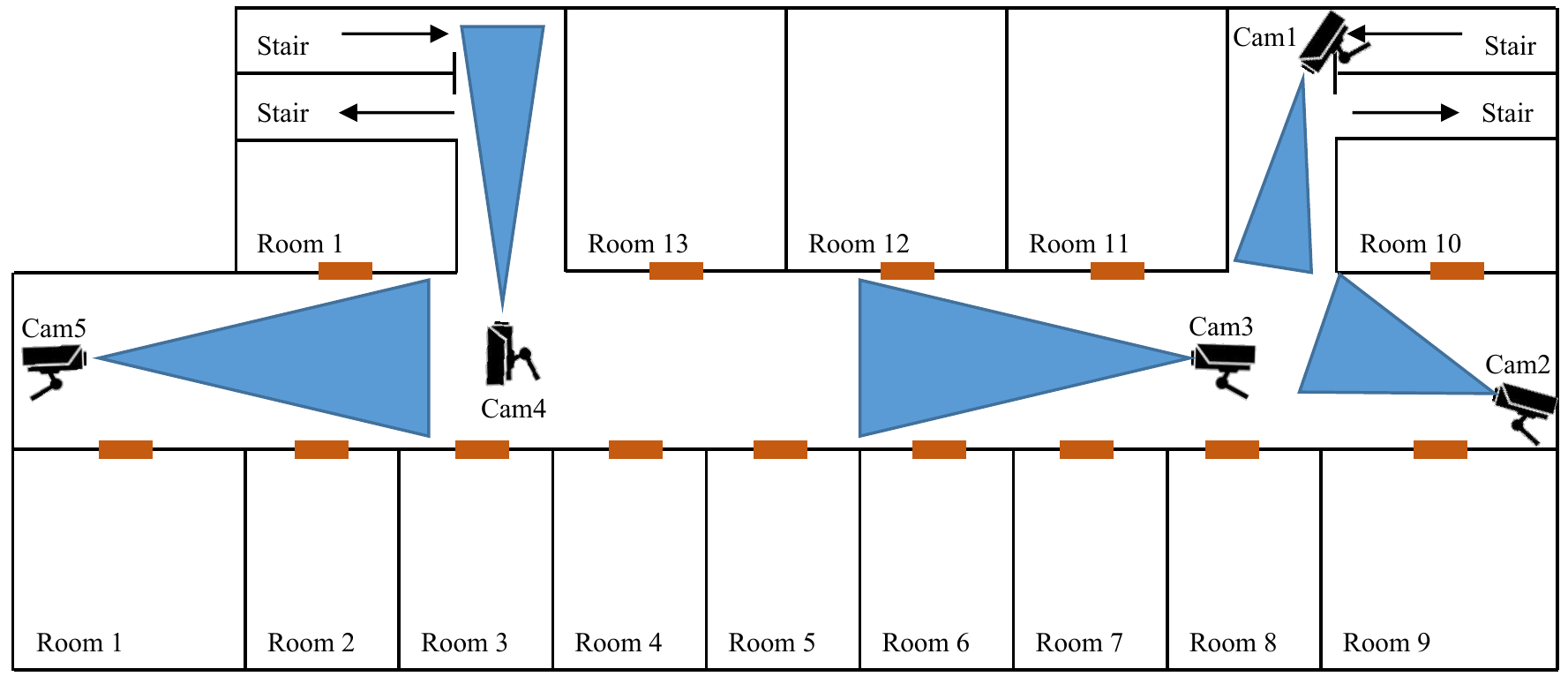}
	\caption{Schematic diagram of five cameras deployed inside the office building}
	\label{fig:experiment_layout}
\end{figure}

Five synchronized 4K UHD ceiling-mounting cameras are installed on the $10^{\text{th}}$ floor of Liuqing Building, Tsinghua University. As shown in Figure \ref{fig:experiment_layout}, the floor area is 1,200 $m^{2}$ and the coverage area of each camera do not overlap with each other. For each video, its frame rate is 25 frames per second, and with High Efficiency Video Coding (HEVC). All the five cameras are linked to a Digital Video Recorder (DVR) with cable connections. In this paper, four business day video data were recorded, and the total length is 480 hours. Undirected camera link graph $\mathcal{G}=(\mathcal{V},\mathcal{E})$ is established, where $\mathcal{V}$ denotes each camera node, and $\mathcal{E}$ is the connectivity between different cameras. Indoor personal movements are treated as transition process among different camera nodes.

\begin{figure}[h]
	\centering
	\includegraphics[scale=0.6]{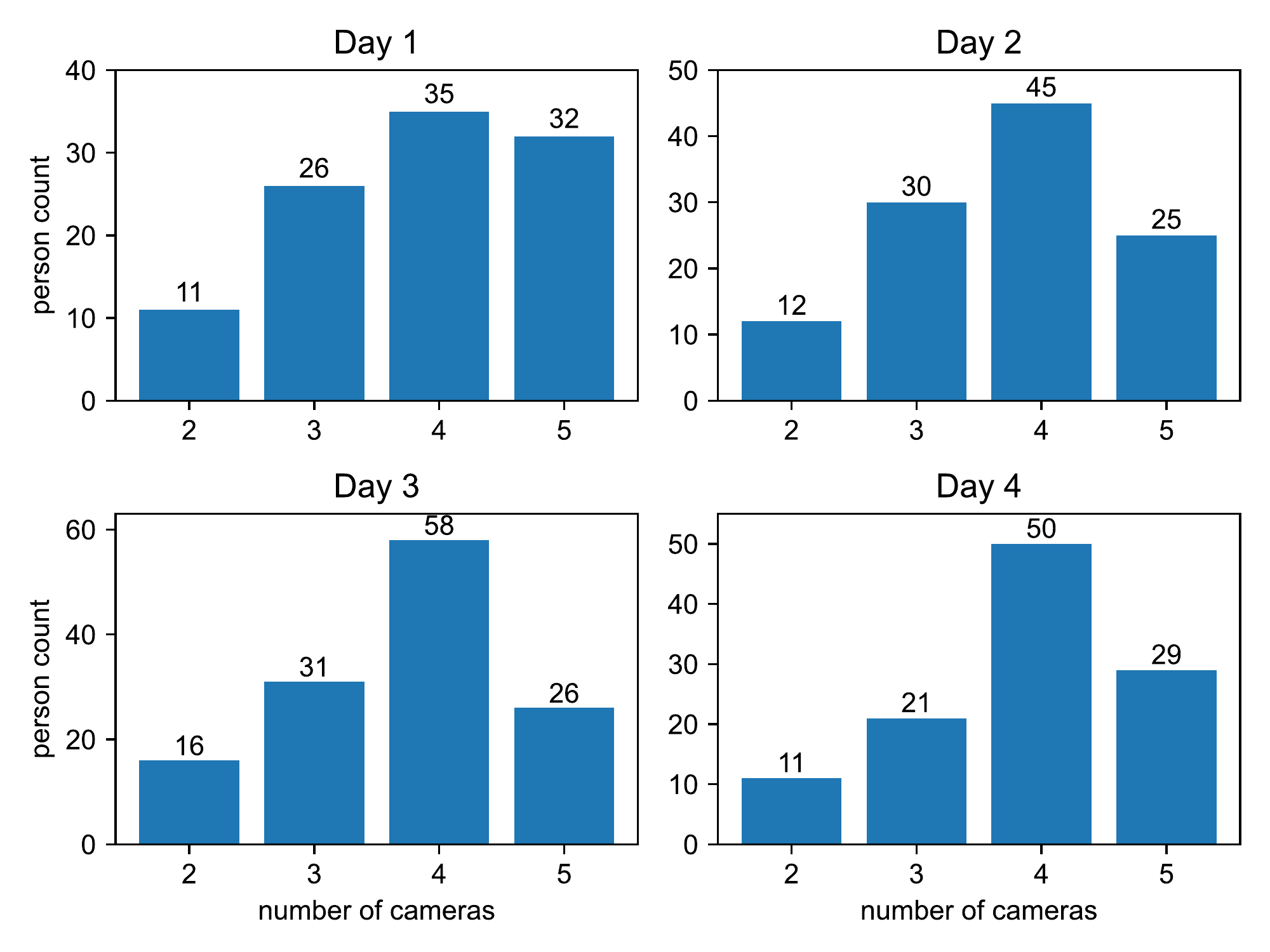}
	\caption{Statistics of the number of single pedestrians appearing in different cameras}
	\label{fig:statistics}
\end{figure}

\begin{figure}[h]
	\centering
	\includegraphics[scale=0.8]{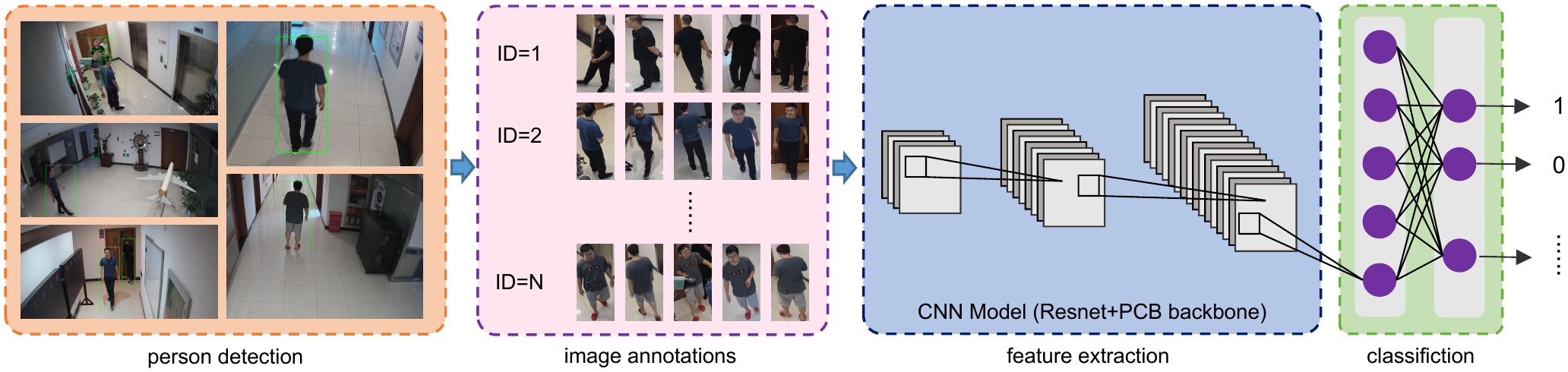}
	\caption{Data preparation and training process}
	\label{fig:reid_CNN}
\end{figure}

This work uses an off-the-shelf person detector \cite{Redmon2018} to get all the bounding boxes of each person who appeared in different videos. To reduce the labor-consumption of image labeling, the detected personal images are cropped from videos in every 25 frames. We devoted to match the same person appeared in different field-of-views, therefore, person who appeared in only one camera is excluded. Video data in the four different business-day are separated, and all the detected images in each day are manually annotated with a globally-unique identifier, i.e. $\mathcal{D}=\{(\mathbf{x_{1}}, y_{1}), (\mathbf{x_{2}}, y_{2}), ..., (\mathbf{x_{N}}, y_{N})\}$ and $N\in\{128, 142, 159, 138\}$, where $\mathbf{x_{i}}$ denotes the detected images of one person from different camera views and $y_{i}$ is the corresponding label. For each day, people was chosen with a random number 80\%, and the corresponding detected images are chosen as training set. The left people and images are as testing set, i.e. $\mathcal{D}=\mathcal{D}_{training} \bigcup \mathcal{D}_{testing}$ where $|\mathcal{D}_{training}|=M$, $|\mathcal{D}_{testing}|=N-M, M \in \{104, 112, 131, 111\}$. Descriptions for the prepared data set are presented in Table \ref{table:surveillance_dataset}. To keep consistency with training data, we random leave one picture out from each training image library as the validation set. In data preparations, each pedestrian appears in at least two cameras, and at most five cameras. The statistics of the number of each single pedestrian appearing in different cameras are shown in Figure \ref{fig:statistics}. For the given training set $\mathcal{D}_{training}$, a supervised deep learning technique, named deep convolutional neural networks (deep ConvNets), is applied for training an appearance feature extractor, which aims for clustering the same person’s images into one group and dividing different person's images into distinct partitions. Data preparation and training process are depicted in Figure \ref{fig:reid_CNN}.

\begin{table}
	\centering
	\caption{Training and testing data set preparation in four days}
	\label{table:surveillance_dataset}
	\begin{tabular}{ccccc}
		\toprule 
		Day&\#Training IDs&\#Training Images&\#Testing IDs&\#Testing Images\\
		\midrule  
		Day 1& 104& 19,313& 24& 4,247\\
		Day 2& 112& 34,403& 30& 6,052\\
		Day 3& 131& 39,150& 28& 4,592\\
		Day 4& 111& 31,072& 27& 4,083\\
		\bottomrule 
	\end{tabular}
\end{table}

\subsection{Appearance based person re-identification}
\subsubsection{Training processes}

Given training set $\mathcal{D}_{training}$, the objective of deep ConvNets is to learn high-level feature embedding function $\hat{y}_{i}=\phi(x_i,\theta)$ to minimize the distance between the estimated value $\hat{y}_{i}$ and the true label $y_{i}$. In this study, we represent the categorical data $\hat{y}_{i}$ and $y_{i}$ as one-hot encoding and the cross entropy between each pair of estimated value and true label is defined as loss function. Therefore, the objective function in the training process is minimize the summation of cross-entropy loss.

\begin{equation}
\label{equ:loss}
\mathcal{L}(\theta)=\min_{\theta} {\sum_{i=1}^{M}l(\phi(x_i;\theta),y_i) }
\end{equation}

\begin{figure}[ht]
	\centering
	\includegraphics[scale=0.6]{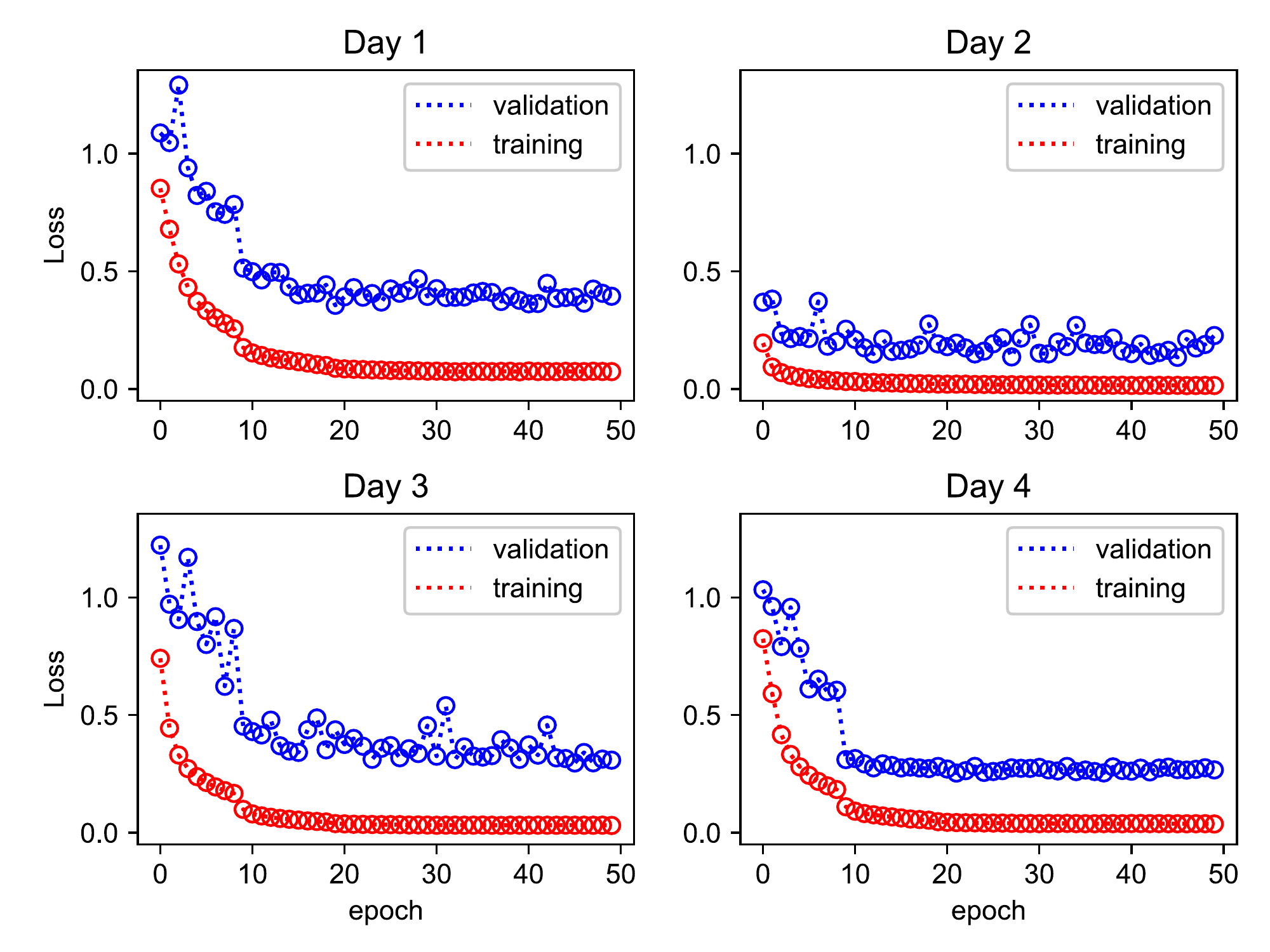}
	\caption{Loss in training and validation processes}
	\label{fig:trainingLoss}
\end{figure}

\begin{figure}[ht]
	\centering
	\includegraphics[scale=0.6]{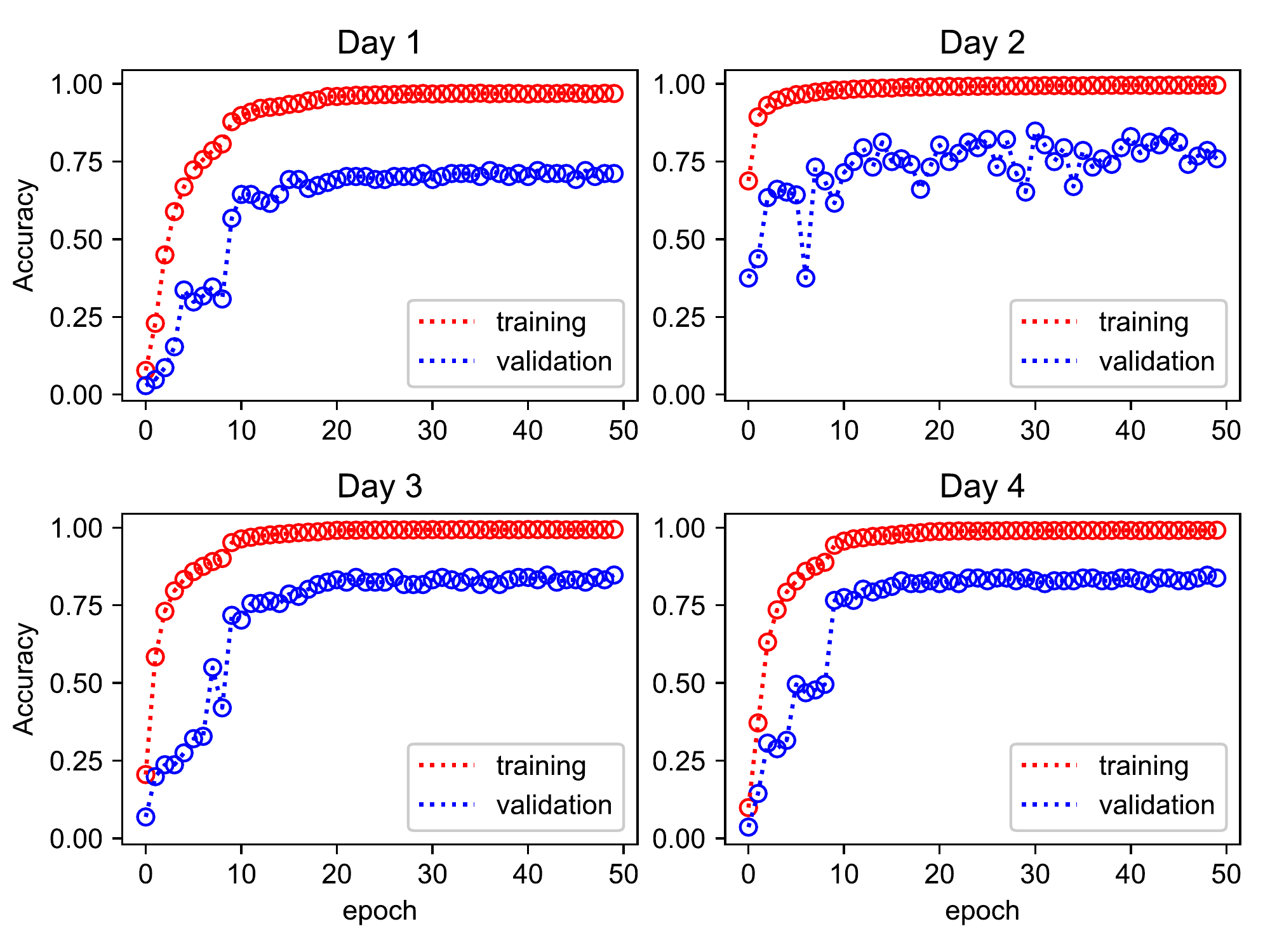}
	\caption{Rank-1 accuracy in training and validation processes}
	\label{fig:trainingTopk}
\end{figure}

In the structure of the deep ConvNets, PCB \cite{sun2019} and ResNet-50 \cite{He2016} are used as pretrained model. In order to perform transfer learning, all the layers are frozen except the output layer, where the total neuron number is modified to equal with the person count $M$ in the training set of different days, i.e. 104, 112, 131, and 111 respectfully. For the training strategies, an stochastic gradient descent approach is applied with back-propagation to update the parameters in each layer of the deep ConvNets. To satisfy the input requirements, all the images in $\mathcal{D}_{training}$ are resized to 384$\times$192$\times$3. The mini-batch size in each iteration is 128 and the total epoch is 50 with learning rate decay in each 10 epochs. Algorithm 1 gives the the training algorithm for indoor person appearance feature extraction. The loss and rank-1 accuracy in training and validation processes are shown in Figure \ref{fig:trainingLoss} and Figure \ref{fig:trainingTopk}.

\begin{algorithm}
	\label{alg: traingAlg}
	\caption{Training algorithm for indoor person appearance feature extraction}
	\KwIn {person count M, person detection results with annotations $\mathcal{D}_{training}$, max number of epochs ($T=50$), number of the deep ConvNets updates per step ($K=|\mathcal{D}_{training}|/batch\_size$)}
	\KwOut{feature embedding function $\phi$ and parameters $\theta$}
	\For{epoch=1,2,...,T}
	{
		\For{step=1,2,...,K}
		{
			1. Sample a mini-batch size of person images in $\mathcal{D}_{training}$;
			
			2. Update $\phi$ and $\theta$ by taking SGD step on mini-batch loss $\mathcal{L}(\theta)$ in (\ref{equ:loss});	
			
			\If{step \% 10 == 0}
			{
				3. decay the learning rate by a factor of 0.1 in every 10 epochs;    
			}

		}
	}

\end{algorithm}

\subsubsection{Testing processes}
In the testing stage, we apply the trained feature embedding function $\phi$ to $\mathcal{D}_{testing}$ and take the output from the fully connected layer as the finally feature embedding for each testing image. Considering PCB and ResNet-50 are used, each feature of testing image is a column vector with size 12,288 (6 parts multiple by 2048, with each part is normalized to 1). The cosine distance between each image pair $\{x_i, x_j\}$ is calculated as appearance similarity between two feature embeddings: 

\begin{equation}
\label{equ:sim1}
sim_{app}(x_i,x_j)=\frac{\phi(x_i;\theta)\cdot\phi(x_j;\theta)}{\left \| \phi(x_i;\theta) \right \|_{2}^{2}  \cdot \left \| \phi(x_i;\theta) \right \|_{2}^{2}} 
\end{equation}

where $x_i, x_j \in \mathcal{D}_{testing}$ and $\phi(x_i;\theta), \phi(x_j;\theta) \in \mathbf{R}^{12288}$, $\left \| \phi(x_i;\theta) \right \|_{2}^{2}=\left \| \phi(x_j;\theta) \right \|_{2}^{2}=6$. 

For each image (query image) in the testing set, we calculate cosine distance to depict the similarity between all the other images (gallery images) and feature similarity matrix are therefore obtained. Meanwhile, to match the indoor person who appeared in different cameras, image pairs of the same individual appeared in the same camera are excluded. Furthermore, by sorting the similarity score and comparing with the image annotations, the Cumulative Matching Characteristic (CMC) curves are obtained, which will be shown by the red curves in Figure \ref{fig:cmc}. 

\begin{figure}[ht]
	\centering
	\includegraphics[scale=0.7]{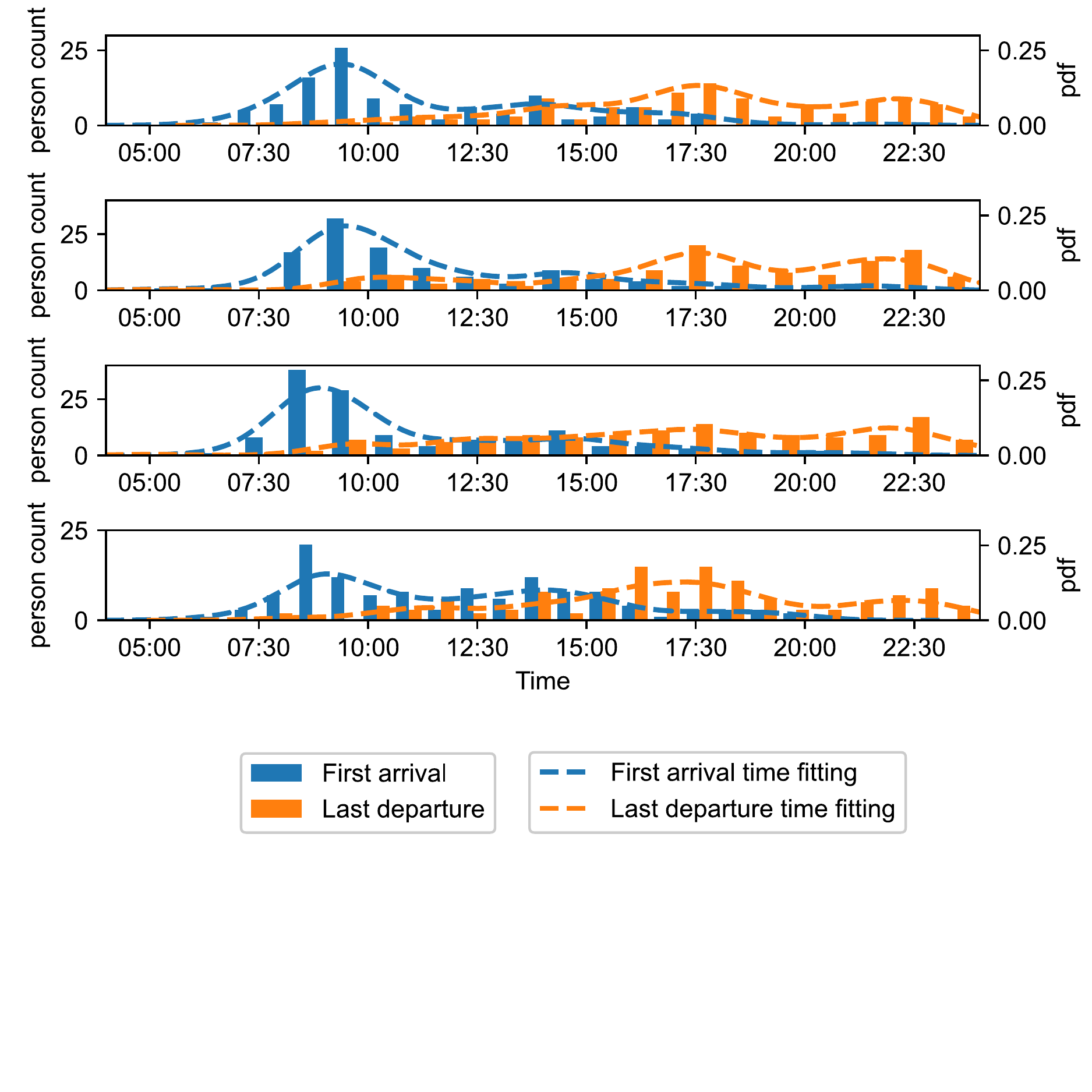}
	\caption{Daily arrive and departure in four different weekdays}
	\label{fig:dayinout}
\end{figure}

\subsection{Transition Pattern Modeling}
\begin{figure}[ht]
	\centering
	\includegraphics[scale=0.5]{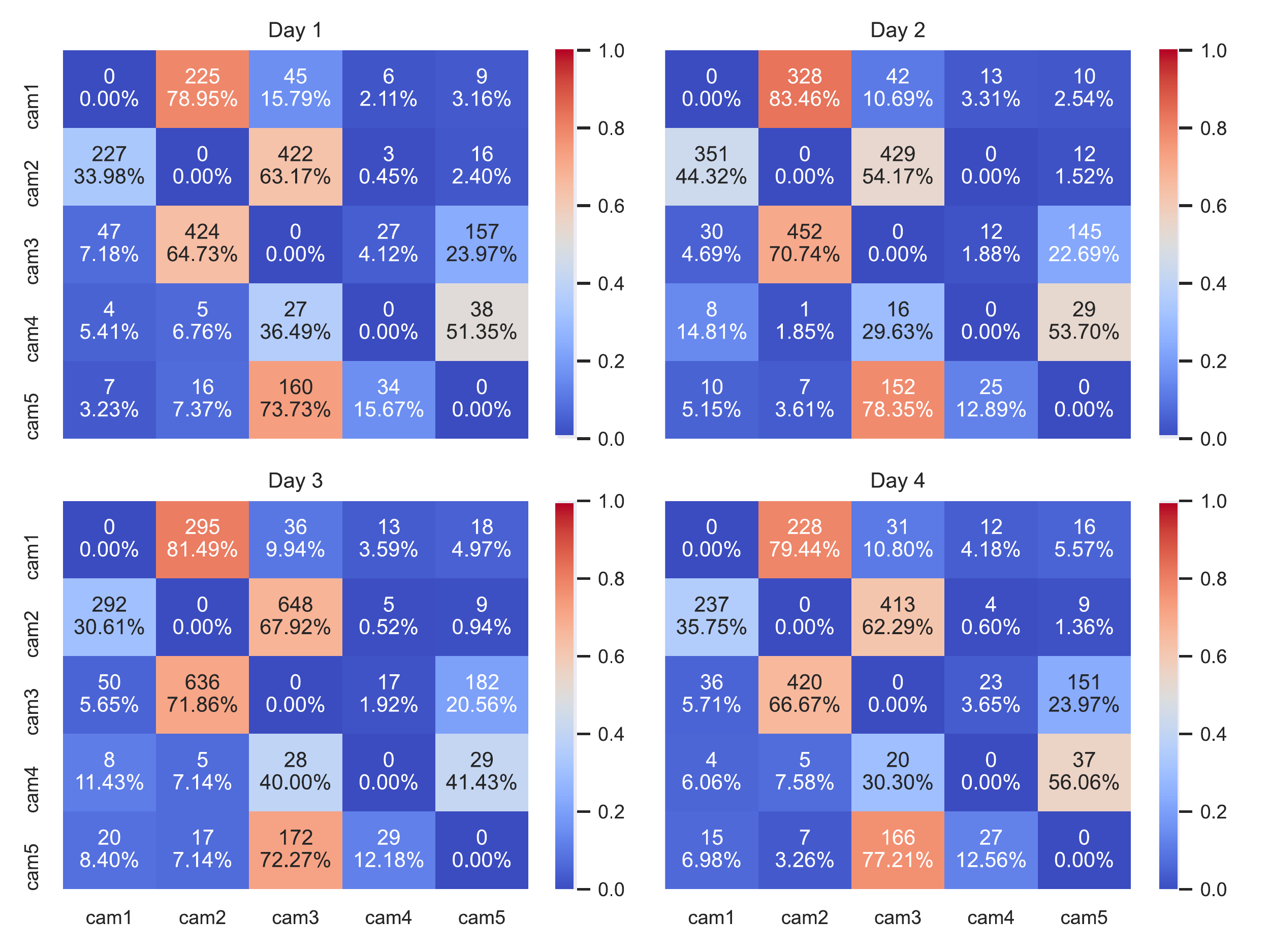}
	\caption{The transition count and state-transition matrix}
	\label{fig:transitionMatrix}
\end{figure}

Considering the dynamic change of personal pose and moving orientations, the appearance feature only is not sufficient to discriminate different occupants. Spatial-temporal constraints are further required to refine the appearance-based person re-identification results. Statistical transition patterns are the repetitive patterns that person takes, which relates to the properties of spatial layout, camera topology, and individual movements. The daily entry and exit count and probability density estimation are shown in Figure \ref{fig:dayinout}. In this study, personal transition between different cameras can be formulated as stochastic matrix. By using the training set, the number and proportion of pedestrians passing between different cameras are calculated. The walk-through selection here is a right stochastic matrix, which means the total of transition probability from camera $i$ to other cameras equals to 1. The first-order state-transition matrix is calculated as follows:

\begin{equation}
\label{equ:stochasticMatrix}
p_{j|i}=\frac{N_{j|i}}{\sum_{j=1}^{\mathcal{N}(i)}{N_{j|i}}}
\end{equation}
s.t. $\sum_{j=1}^{\mathcal{N}(i)}{p_{j|i}}=1$, where $N_{j|i}$, $p_{j|i}$ denotes the number and proportion of person transition from camera $i$ to camera $j$ respectfully. $\mathcal{N}(i)$ means the neighbor cameras of camera $i$, which can be obtain from undirected camera link graph $\mathcal{G}$. The transition count and state-transition matrix between among different cameras are therefore calculated and shown in Figure \ref{fig:transitionMatrix}.

\subsection{Results}
Coupling appearance feature with the state-transition matrix, similarity between the query image $x_i$ and gallery image $x_j$ is further formulated as: 

\begin{equation}
\label{equ:sim2_1}
sim(x_i,x_j)=sim_{app}(x_i, x_j) \cdot p_{j|i}
\end{equation}

\begin{equation}
\label{equ:sim2_2}
sim(x_i,x_j)=\frac{\phi(x_i;\theta)\cdot\phi(x_j;\theta)}{\left \| \phi(x_i;\theta) \right \|_{2}^{2}  \cdot \left \| \phi(x_i;\theta) \right \|_{2}^{2}}  \cdot \frac{N_{j|i}}{\sum_{j=1}^{\mathcal{N}(i)}{N_{j|i}}}
\end{equation}

\begin{figure}[ht]
	\centering
	\includegraphics[scale=0.6]{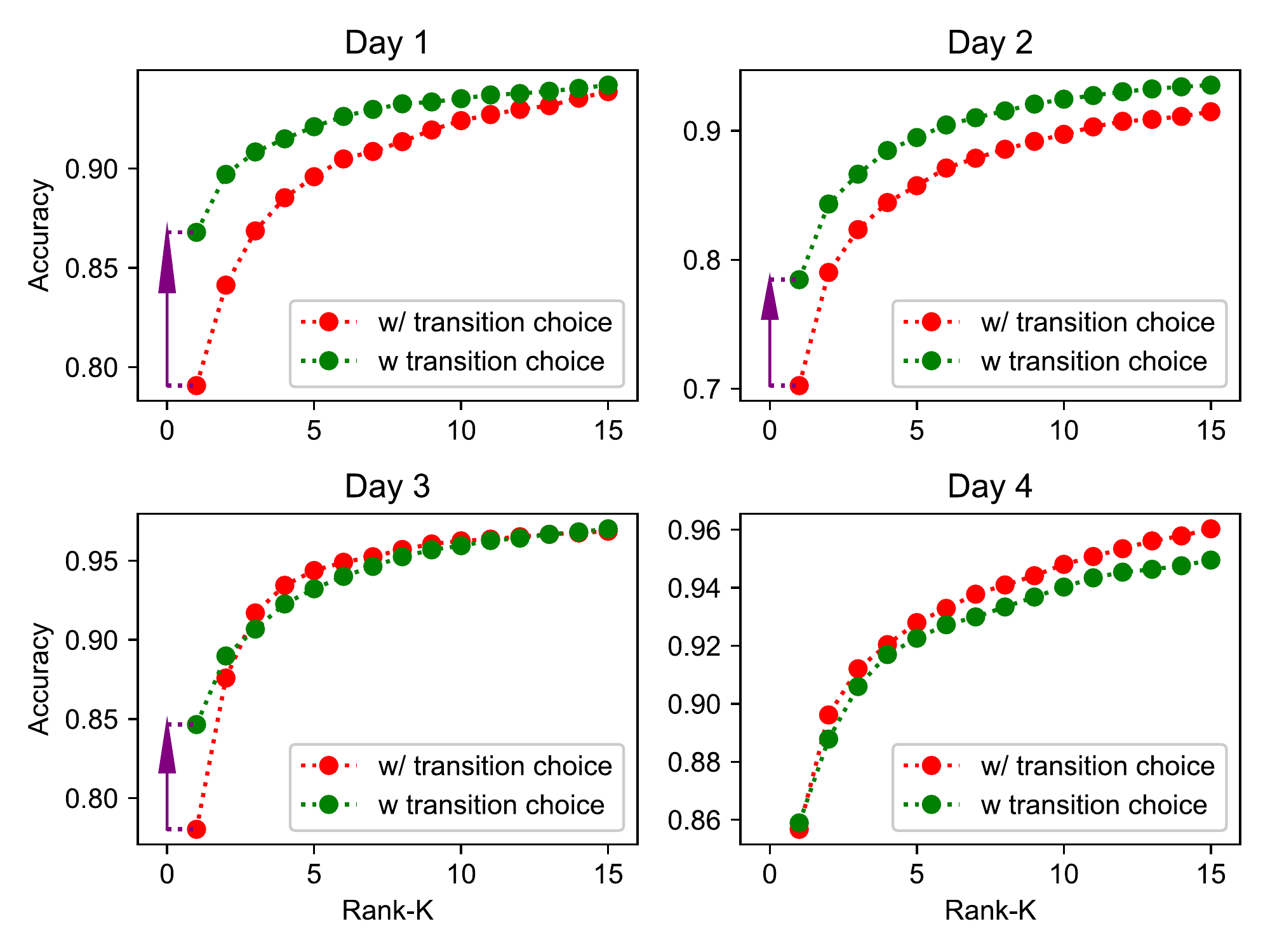}
	\caption{Rank-K accuracy improvements in person re-identification results}
	\label{fig:cmc}
\end{figure}

Table \ref{table:cmcDay1},\ref{table:cmcDay2},\ref{table:cmcDay3},\ref{table:cmcDay4} demonstrates several representative values in the Cumulative Matching Characteristic curve and the improvements when adding state-transition matrix to sole appearance feature. Experimental results show that the Rank-1 accuracy has been improved by 7.72\%, 8.21\%, 6.62\% and 0.22\% respectfully. The average improvement is 5.69\%. The full ranking results in the Cumulative Matching Characteristic curve is shown in Figure \ref{fig:cmc}, where the red curves shows the sole-appearance feature used, and the green curves represent the improvements after adding spatial constrains. The purple arrows demonstrate the corresponding rank-1 improvements.

\begin{table}
	\centering
	\caption{The value of CMC changes before and after adding state-transition matrix (Day 1)}
	\label{table:cmcDay1}
	\begin{tabular}{ccccc}
		\toprule 
		Day 1&Rank-1&Rank-5&Rank-10&Rank-15\\
		\midrule  
		Appearance&	0.7907&	0.8959&	0.9242&	0.9388\\
		w/ transition choice&	0.8679 (↑)&	0.9211 (↑)&	0.9352 (↑)&	0.9421 (↑) \\
		\bottomrule 
	\end{tabular}
\end{table}

\begin{table}
	\centering
	\caption{The value of CMC changes before and after adding state-transition matrix (Day 2)}
	\label{table:cmcDay2}
	\begin{tabular}{ccccc}
		\toprule 
		Day 2&Rank-1&Rank-5&Rank-10&Rank-15\\
		\midrule  
		Appearance&	0.7024&	0.8574&	0.8972&	0.9147 \\
		w/ transition choice&	0.7845 (↑)&	0.8947 (↑)&	0.9245 (↑)&	0.9354 (↑) \\
		\bottomrule 
	\end{tabular}
\end{table}

\begin{table}
	\centering
	\caption{The value of CMC changes before and after adding state-transition matrix (Day 3)}
	\label{table:cmcDay3}
	\begin{tabular}{ccccc}
		\toprule 
		Day 3&Rank-1&Rank-5&Rank-10&Rank-15\\
		\midrule  
		Appearance&	0.7803&	0.9438&	0.9625&	0.9686 \\
		w/ transition choice&	0.8465 (↑)&	0.9323&	0.9595&	0.9702 \\
		\bottomrule 
	\end{tabular}
\end{table}

\begin{table}
	\centering
	\caption{The value of CMC changes before and after adding state-transition matrix (Day 4)}
	\label{table:cmcDay4}
	\begin{tabular}{ccccc}
		\toprule 
		Day 4&Rank-1&Rank-5&Rank-10&Rank-15\\
		\midrule  
		Appearance&	0.8567&	0.928&	0.9481&	0.9603 \\
		w/ transition choice&	0.8589 (↑)&	0.9226&	0.9402&	0.9495 \\
		\bottomrule 
	\end{tabular}
\end{table}

\begin{figure}[H]
	\centering
	\includegraphics[scale=0.5]{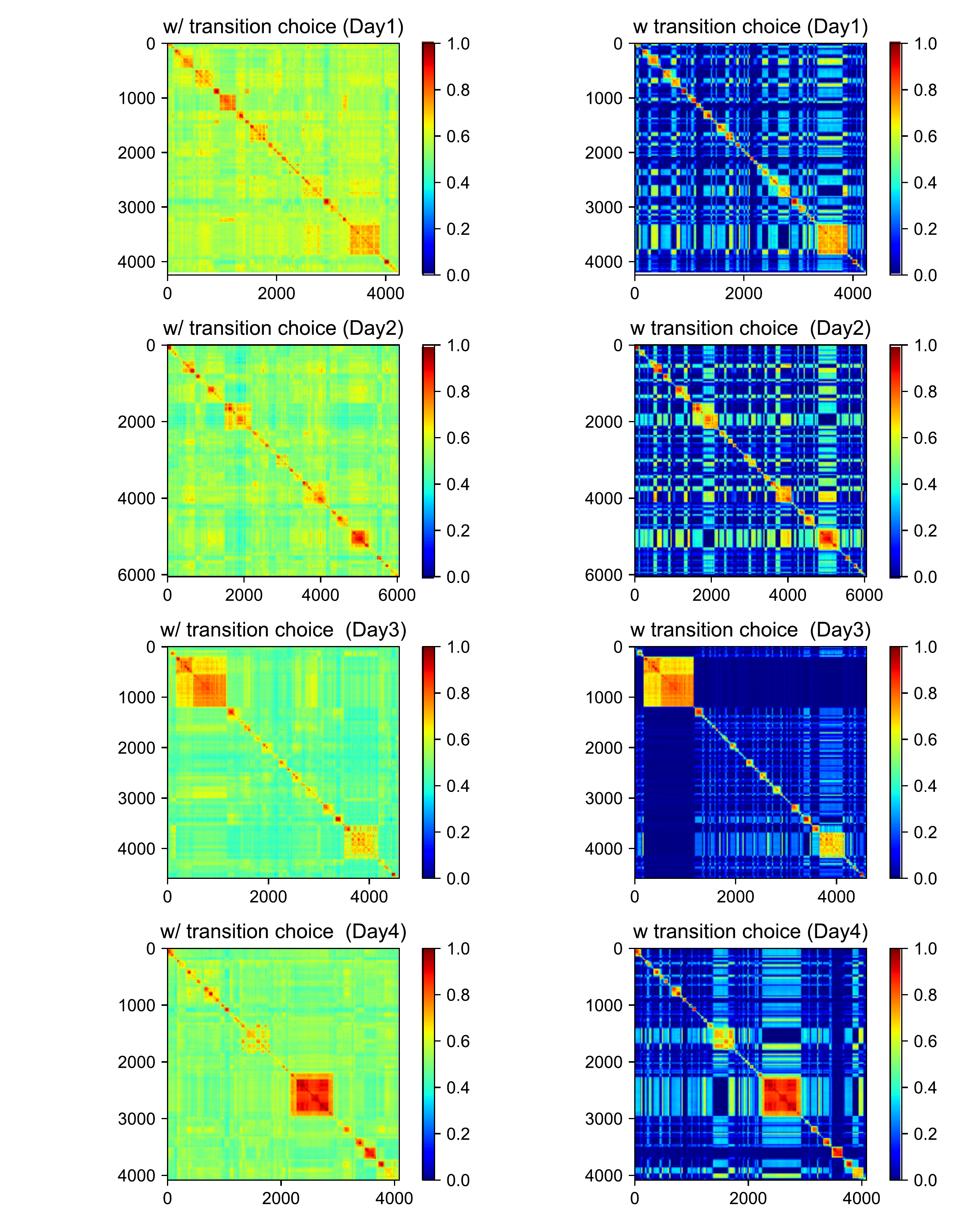}
	\caption{Comparison between similarities when adding transition choice}
	\label{fig:heatmap}
\end{figure}

We normalized the final pedestrian similarity matrix, and visualized the changes in the appearance similarity heat map after transition choice is added. As shown Figure \ref{fig:heatmap}, the similairty matrix is symmetric, where the image count equals with the total testing images number as shown in Table \ref{table:surveillance_dataset}, i.e., 4247, 6052, 4592, and 4083 respectfully. The value of each pixel in the similarity heat map indicates the level of final similarity score between query image $x_i$ and gallery image $x_j$, where 1 indicate the same person, and 0 is the different. Each person has multiple images under different cameras, and the small rectangles in the main diagonal area express the similarity of the same pedestrians appeared in other different cameras. As can be seen from the figure in the right-hand side, by adding state transition constraints, the ranking algorithm actually do a step further for filtering and suppressing the matching score when people walking among different cameras with low probabilities. Relatively, the similarity of the same pedestrian under different cameras are improved.

A specific person re-identification example is demonstrated. Given a probe image, the ranking algorithm retrieves the best 10 matches from the gallery, i.e. the personnel images captured by different cameras. The digital number in green color representdongs the correct matching, and the red is the wrong matching. From the figure we can see that more correct match is retrieved when transition choice is combined with appearance feature embeddings.

\section{Discussion}
In this paper, a multi-camera system, which consists of five non-overlapping surveillance cameras, is deployed inside an office building. A graph-based camera linkage is further established to extend the range of the pedestrian movement monitoring. For matching the person images captured by different cameras, a deep convolutional neural network is trained to extract robust appearance feature embeddings and similarity comparison is therefore applied. The ratio of the correct matching in rank-1 accuracy over Cumulative Matching Characteristics curve reaches to 78.25\% on average. By leveraging statistical transition patterns, the false match with low similarity score in the ranking list is further filtered out. Coupling the state-transition matrix constraints with appearance feature similarities, the rank-1 accuracy increased by 5.69\% on average, which shows potential for further improvements when more video data and transition choice are accumulated.

\section{Conclusions}
To compensate for the insufficient coverage of single camera, this study demonstrates a multi-camera system to achieve a wider range of scene capturing, which can be applied to depict a larger-scale indoor personal movement monitoring. More than 120 pedestrians moving inside an building in each 4 different weekdays are monitored by the proposed system.

Meanwhile, there is no additional type of sensors required, and does not need active cooperation from indoor occupants. The appearance feature embeddings are associated with the state-transition matrix among multiple non-overlapping cameras to improve the accuracy of pedestrian re-identification. By adding the selection of transition choice to the appearance feature embeddings, the rank-1 accuracy of cross-camera pedestrian matching is improved to 83.94\%, which can be treated as a baseline for further improvements. At the same time, these cameras are installed in public indoor area and do not rely on high-definition human faces, which maximize privacy protection. 

Our future research is devoted to unveiling the transition time distribution when pedestrians passing among different cameras. By sorting the similarity score by time sequence, we aim to further restoring the indoor personal trajectory in multi-camera circumstance. 

\section*{Acknowledgments}
This work is supported by National Key Research and Development Project
of China (No.2017YFC0704100, and No.2017YFC0803300)  and National Natural Science
Foundation of China (Grant U2033206).


\bibliography{mybibfile}

\begin{thebibliography}{10}
\expandafter\ifx\csname url\endcsname\relax
  \def\url#1{\texttt{#1}}\fi
\expandafter\ifx\csname urlprefix\endcsname\relax\def\urlprefix{URL }\fi
\expandafter\ifx\csname href\endcsname\relax
  \def\href#1#2{#2} \def\path#1{#1}\fi

\bibitem{Sun2020}
K.~Sun, Q.~Zhao, J.~Zou, {A review of building occupancy measurement systems},
  Energy and Buildings 216 (2020) 109965.
\newblock \href {http://dx.doi.org/10.1016/j.enbuild.2020.109965}
  {\path{doi:10.1016/j.enbuild.2020.109965}}.

\bibitem{Saha2019}
H.~Saha, A.~R. Florita, G.~P. Henze, S.~Sarkar, {Occupancy sensing in
  buildings: A review of data analytics approaches}, Energy and Buildings
  188-189 (2019) 278--285.
\newblock \href {http://dx.doi.org/10.1016/j.enbuild.2019.02.030}
  {\path{doi:10.1016/j.enbuild.2019.02.030}}.

\bibitem{Rueda2020}
L.~Rueda, K.~Agbossou, A.~Cardenas, N.~Henao, S.~Kelouwani, {A comprehensive
  review of approaches to building occupancy detection}, Building and
  Environment 180 (2020) 106966.
\newblock \href {http://dx.doi.org/10.1016/j.buildenv.2020.106966}
  {\path{doi:10.1016/j.buildenv.2020.106966}}.

\bibitem{Benezeth2011}
Y.~Benezeth, H.~Laurent, B.~Emile, C.~Rosenberger, {Towards a sensor for
  detecting human presence and characterizing activity}, Energy and Buildings
  43~(2-3) (2011) 305--314.
\newblock \href {http://dx.doi.org/10.1016/j.enbuild.2010.09.014}
  {\path{doi:10.1016/j.enbuild.2010.09.014}}.

\bibitem{Yang2016}
J.~Yang, M.~Santamouris, S.~E. Lee, {Review of occupancy sensing systems and
  occupancy modeling methodologies for the application in institutional
  buildings}, Energy and Buildings 121 (2016) 344--349.
\newblock \href {http://dx.doi.org/10.1016/j.enbuild.2015.12.019}
  {\path{doi:10.1016/j.enbuild.2015.12.019}}.

\bibitem{Wang2013}
X.~Wang, {Intelligent multi-camera video surveillance: A review}, Pattern
  Recognition Letters 34~(1) (2013) 3--19.
\newblock \href {http://dx.doi.org/10.1016/j.patrec.2012.07.005}
  {\path{doi:10.1016/j.patrec.2012.07.005}}.

\bibitem{Petersen2016}
S.~Petersen, T.~H. Pedersen, K.~U. Nielsen, M.~D. Knudsen, {Establishing an
  image-based ground truth for validation of sensor data-based room occupancy
  detection}, Energy and Buildings 130 (2016) 787--793.
\newblock \href {http://dx.doi.org/10.1016/j.enbuild.2016.09.009}
  {\path{doi:10.1016/j.enbuild.2016.09.009}}.

\bibitem{Zou2017}
J.~Zou, Q.~Zhao, W.~Yang, F.~Wang, {Occupancy detection in the office by
  analyzing surveillance videos and its application to building energy
  conservation}, Energy and Buildings 152 (2017) 385--398.
\newblock \href {http://dx.doi.org/10.1016/j.enbuild.2017.07.064}
  {\path{doi:10.1016/j.enbuild.2017.07.064}}.

\bibitem{Chen2009}
L.~Chen, F.~Chen, X.~Guan, {A video-based indoor occupant detection and
  localization algorithm for smart buildings}, Lecture Notes in Computer
  Science (including subseries Lecture Notes in Artificial Intelligence and
  Lecture Notes in Bioinformatics) 5754 LNCS (2009) 565--573.
\newblock \href {http://dx.doi.org/10.1007/978-3-642-04070-2_62}
  {\path{doi:10.1007/978-3-642-04070-2_62}}.

\bibitem{Gao2016}
C.~Gao, P.~Li, Y.~Zhang, J.~Liu, L.~Wang, {People counting based on head
  detection combining Adaboost and CNN in crowded surveillance environment},
  Neurocomputing 208 (2016) 108--116.
\newblock \href {http://dx.doi.org/10.1016/j.neucom.2016.01.097}
  {\path{doi:10.1016/j.neucom.2016.01.097}}.

\bibitem{Kuipers2014}
M.~Kuipers, A.~Tom{\'{e}}, T.~Pinheiro, M.~Nunes, T.~Heitor, {Building
  space-use analysis system - A multi location/multi sensor platform},
  Automation in Construction 47 (2014) 10--23.
\newblock \href {http://dx.doi.org/10.1016/j.autcon.2014.07.001}
  {\path{doi:10.1016/j.autcon.2014.07.001}}.

\bibitem{Shih2014}
H.~C. Shih, {A robust occupancy detection and tracking algorithm for the
  automatic monitoring and commissioning of a building}, Energy and Buildings
  77~(May 2013) (2014) 270--280.
\newblock \href {http://dx.doi.org/10.1016/j.enbuild.2014.03.069}
  {\path{doi:10.1016/j.enbuild.2014.03.069}}.

\bibitem{ding2012collaborative}
C.~Ding, B.~Song, A.~Morye, J.~A. Farrell, A.~K. Roy-Chowdhury, Collaborative
  sensing in a distributed ptz camera network, IEEE Transactions on Image
  Processing 21~(7) (2012) 3282--3295.
\newblock \href {http://dx.doi.org/10.1109/TIP.2012.2188806}
  {\path{doi:10.1109/TIP.2012.2188806}}.

\bibitem{Liu2013}
D.~Liu, X.~Guan, Y.~Du, Q.~Zhao, {Measuring indoor occupancy in intelligent
  buildings using the fusion of vision sensors}, Measurement Science and
  Technology 24~(7).
\newblock \href {http://dx.doi.org/10.1088/0957-0233/24/7/074023}
  {\path{doi:10.1088/0957-0233/24/7/074023}}.

\bibitem{Camps2017}
O.~Camps, M.~Gou, T.~Hebble, S.~Karanam, O.~Lehmann, Y.~Li, R.~J. Radke, Z.~Wu,
  F.~Xiong, {From the Lab to the Real World: Re-identification in an Airport
  Camera Network}, IEEE Transactions on Circuits and Systems for Video
  Technology 27~(3) (2017) 540--553.
\newblock \href {http://dx.doi.org/10.1109/TCSVT.2016.2556538}
  {\path{doi:10.1109/TCSVT.2016.2556538}}.

\bibitem{Figueira2015}
D.~Figueira, M.~Taiana, A.~Nambiar, J.~Nascimento, A.~Bernardino, {The HDA+
  data set for research on fully automated re-identification systems}, Lecture
  Notes in Computer Science (including subseries Lecture Notes in Artificial
  Intelligence and Lecture Notes in Bioinformatics) 8927 (2015) 241--255.
\newblock \href {http://dx.doi.org/10.1007/978-3-319-16199-0_17}
  {\path{doi:10.1007/978-3-319-16199-0_17}}.

\bibitem{Bialkowski2012}
A.~Bialkowski, S.~Denman, S.~Sridharan, C.~Fookes, P.~Lucey, {A database for
  person re-identification in multi-camera surveillance networks}, 2012
  International Conference on Digital Image Computing Techniques and
  Applications, DICTA 2012\href {http://dx.doi.org/10.1109/DICTA.2012.6411689}
  {\path{doi:10.1109/DICTA.2012.6411689}}.

\bibitem{Marroquin2019}
R.~Marroquin, J.~Dubois, C.~Nicolle, {WiseNET: An indoor multi-camera
  multi-space dataset with contextual information and annotations for people
  detection and tracking}, Data in Brief 27 (2019) 104654.
\newblock \href {http://dx.doi.org/10.1016/j.dib.2019.104654}
  {\path{doi:10.1016/j.dib.2019.104654}}.

\bibitem{Styles2020}
O.~Styles, T.~Guha, V.~Sanchez, A.~Kot, {Multi-camera trajectory forecasting:
  Pedestrian trajectory prediction in a network of cameras}, IEEE Computer
  Society Conference on Computer Vision and Pattern Recognition Workshops
  2020-June (2020) 4379--4382.
\newblock \href {http://arxiv.org/abs/2005.00282} {\path{arXiv:2005.00282}},
  \href {http://dx.doi.org/10.1109/CVPRW50498.2020.00516}
  {\path{doi:10.1109/CVPRW50498.2020.00516}}.

\bibitem{Bedagkar-Gala2014}
A.~Bedagkar-Gala, S.~K. Shah, {A survey of approaches and trends in person
  re-identification}, Image and Vision Computing 32~(4) (2014) 270--286.
\newblock \href {http://dx.doi.org/10.1016/j.imavis.2014.02.001}
  {\path{doi:10.1016/j.imavis.2014.02.001}}.

\bibitem{Zheng2016}
L.~Zheng, Y.~Yang, A.~G. Hauptmann,
  \href{http://arxiv.org/abs/1610.02984}{{Person Re-identification: Past,
  Present and Future}} 14~(8) (2016) 1--20.
\newblock \href {http://arxiv.org/abs/1610.02984} {\path{arXiv:1610.02984}}.
\newline\urlprefix\url{http://arxiv.org/abs/1610.02984}

\bibitem{Karanam2019}
S.~Karanam, M.~Gou, Z.~Wu, A.~Rates-Borras, O.~Camps, R.~J. Radke, {A
  Systematic Evaluation and Benchmark for Person Re-Identification: Features,
  Metrics, and Datasets}, IEEE Transactions on Pattern Analysis and Machine
  Intelligence 41~(3) (2019) 523--536.
\newblock \href {http://arxiv.org/abs/1605.09653} {\path{arXiv:1605.09653}},
  \href {http://dx.doi.org/10.1109/TPAMI.2018.2807450}
  {\path{doi:10.1109/TPAMI.2018.2807450}}.

\bibitem{Leng2020}
Q.~Leng, M.~Ye, Q.~Tian, {A Survey of Open-World Person Re-Identification},
  IEEE Transactions on Circuits and Systems for Video Technology 30~(4) (2020)
  1092--1108.
\newblock \href {http://dx.doi.org/10.1109/TCSVT.2019.2898940}
  {\path{doi:10.1109/TCSVT.2019.2898940}}.

\bibitem{Ye2020}
M.~Ye, J.~Shen, G.~Lin, T.~Xiang, L.~Shao, S.~C.~H. Hoi,
  \href{http://arxiv.org/abs/2001.04193}{{Deep Learning for Person
  Re-identification: A Survey and Outlook}} (2020) 1--20\href
  {http://arxiv.org/abs/2001.04193} {\path{arXiv:2001.04193}}.
\newline\urlprefix\url{http://arxiv.org/abs/2001.04193}

\bibitem{Chen2011}
K.~W. Chen, C.~C. Lai, P.~J. Lee, C.~S. Chen, Y.~P. Hung, {Adaptive learning
  for target tracking and true linking discovering across multiple
  non-overlapping cameras}, IEEE Transactions on Multimedia 13~(4) (2011)
  625--638.
\newblock \href {http://dx.doi.org/10.1109/TMM.2011.2131639}
  {\path{doi:10.1109/TMM.2011.2131639}}.

\bibitem{Chen2014}
X.~Chen, K.~Huang, T.~Tan, {Object tracking across non-overlapping views by
  learning inter-camera transfer models}, Pattern Recognition 47~(3) (2014)
  1126--1137.
\newblock \href {http://dx.doi.org/10.1016/j.patcog.2013.06.011}
  {\path{doi:10.1016/j.patcog.2013.06.011}}.

\bibitem{Javed2008}
O.~Javed, K.~Shafique, Z.~Rasheed, M.~Shah, {Modeling inter-camera space-time
  and appearance relationships for tracking across non-overlapping views},
  Computer Vision and Image Understanding 109~(2) (2008) 146--162.
\newblock \href {http://dx.doi.org/10.1016/j.cviu.2007.01.003}
  {\path{doi:10.1016/j.cviu.2007.01.003}}.

\bibitem{Wang2019}
G.~Wang, J.~Lai, P.~Huang, X.~Xie, {Spatial-temporal person re-identification},
  33rd AAAI Conference on Artificial Intelligence, AAAI 2019, 31st Innovative
  Applications of Artificial Intelligence Conference, IAAI 2019 and the 9th
  AAAI Symposium on Educational Advances in Artificial Intelligence, EAAI 2019
  2 (2019) 8933--8940.
\newblock \href {http://arxiv.org/abs/1812.03282} {\path{arXiv:1812.03282}},
  \href {http://dx.doi.org/10.1609/aaai.v33i01.33018933}
  {\path{doi:10.1609/aaai.v33i01.33018933}}.

\bibitem{Redmon2018}
J.~Redmon, A.~Farhadi, {YOLOv3: An incremental improvement}, arXiv\href
  {http://arxiv.org/abs/1804.02767} {\path{arXiv:1804.02767}}.

\bibitem{sun2019}
Y.~Sun, L.~Zheng, Y.~Li, Y.~Yang, Q.~Tian, S.~Wang, {Learning Part-based
  Convolutional Features for Person Re-identification}, IEEE Transactions on
  Pattern Analysis and Machine Intelligence PP~(c) (2019) 1--1.
\newblock \href {http://dx.doi.org/10.1109/tpami.2019.2938523}
  {\path{doi:10.1109/tpami.2019.2938523}}.

\bibitem{He2016}
K.~He, X.~Zhang, S.~Ren, J.~Sun, {Deep residual learning for image
  recognition}, Proceedings of the IEEE Computer Society Conference on Computer
  Vision and Pattern Recognition 2016-December (2016) 770--778.
\newblock \href {http://arxiv.org/abs/1512.03385} {\path{arXiv:1512.03385}},
  \href {http://dx.doi.org/10.1109/CVPR.2016.90}
  {\path{doi:10.1109/CVPR.2016.90}}.

\end{thebibliography}

\end{document}